\documentclass{article}

\usepackage{microtype}
\usepackage{subfigure}
\usepackage{amsmath,amssymb}
\DeclareMathOperator{\E}{\mathbb{E}}
\newcommand{\cut}[1]{}

\usepackage{booktabs} 
\usepackage{microtype}      
\usepackage[utf8]{inputenc}
\usepackage[english]{babel}
\newcommand{\xhdr}[1]{{\noindent\bfseries #1}.}
\usepackage{xcolor}
\usepackage{graphicx}
\usepackage{hyperref}

\usepackage[accepted]{icml2019}

\icmltitlerunning{Improving Exploration in Soft-Actor-Critic with Normalizing Flows Policies}

\begin{document}

\twocolumn[
\icmltitle{Improving Exploration in Soft-Actor-Critic with Normalizing Flows Policies}



\icmlsetsymbol{equal}{*}

\begin{icmlauthorlist}
\icmlauthor{Patrick Nadeem Ward $^*$}{mcgill,mila}
\icmlauthor{Ariella Smofsky $^*$}{mcgill,mila}
\icmlauthor{Avishek Joey Bose}{mcgill,mila}
\end{icmlauthorlist}

\icmlaffiliation{mcgill}{McGill University}
\icmlaffiliation{mila}{Mila}

\icmlcorrespondingauthor{Patrick Nadeem Ward}{patrick.ward@mail.mcgill.ca}
\icmlcorrespondingauthor{Ariella Smofsky}{ariella.smofsky@mail.mcgill.ca}

\icmlkeywords{Machine Learning, ICML}

\vskip 0.3in
]



\printAffiliationsAndNotice{\icmlEqualContribution} 

\begin{abstract}
Deep Reinforcement Learning (DRL) algorithms for continuous action spaces are known to be brittle toward hyperparameters as well as \cut{being}sample inefficient. Soft Actor Critic (SAC) proposes an off-policy deep actor critic algorithm within the maximum entropy RL framework which offers greater stability and empirical gains.  The choice of policy distribution, a factored Gaussian, is motivated by \cut{chosen due}its easy re-parametrization rather than its modeling power. We introduce Normalizing Flow policies within the SAC framework that learn more expressive classes of policies than simple factored Gaussians. \cut{We also present a series of stabilization tricks that enable effective training of these policies in the RL setting.}We show empirically on continuous grid world tasks that our approach increases stability and is better suited to difficult exploration in sparse reward settings.
\end{abstract}

\section{Introduction}
Policy Gradient methods have been the defacto gold standard for scaling up Reinforcement Learning with non-linear function approximation.\cut{In many of these tasks the action spaces can either be discrete or continuous; the latter often being more challenging.} In continuous control tasks stochastic policies have the added benefit of permitting on-policy exploration and learning from sampled experience in an off-policy manner \cite{heess2015learning}. Soft Actor Critic presents a stable model free DRL algorithm which augments the standard maximum reward
Reinforcement Learning objective with an entropy maximization term \cite{haarnoja2018soft}. SAC allows for the learning of policies that maximize both expected reward and policy entropy which aids in stability, exploration and robustness \cite{ziebart2008maximum,ziebart2010modeling}. Nevertheless, most stochastic policies are Gaussian policies that make use of the reparameterization trick in generative modeling to directly optimize distributional parameters via the pathwise gradient estimator \cite{schulman2015gradient}. While computationally attractive, Gaussian policies have limited modeling expressivity i.e. only policies that are diagonal Gaussians. In many learning tasks the choice of target policy can be a limiting assumption. For instance, in real world robotics tasks where actions may correspond to bounded joint angles due to physical constraints, a Beta policy has been shown to converge faster and to better policies \cite{chou2017improving}.\cut{Within, the maximum entropy framework the expressiveness of the stochastic policy has a direct correlation with the exploratory behavior through the policy's entropy. Thus, when needed, more expressive classes of policies have the ability to model complex environments while exploring efficiently even in sparse reward settings.}

In this paper, we analyze the utility of modeling stochastic policies using different probability distributions for continuous control tasks. In most cases the form of the optimal policy is not known \textit{a priori} and thus picking an appropriate distribution for modeling stochastic policies can be difficult without domain specific knowledge. We propose to bypass this modeling bottleneck by introducing Normalizing Flow policies that define a sequence of invertible transformations that map a sample from a simple density to a more complex density via the change of variable formula for probability distributions \cite{rezende2015variational}. In short, these policies can learn arbitrarily complex distributions but still retain the attractive properties of sampling from a simple distribution such as a reparameterized Gaussian and having a defined density making it amenable to the maximum entropy RL framework. While past work applies Normalizing flows for on-policy learning \cite{tang2018boosting}, we instead apply Normalizing Flows to Soft Actor Critic. We argue the use of Normalizing Flows is a natural choice as the squashing function used in SAC to enforce action bounds can be viewed as a simple one layer flow. Extending SAC with multiple and more complex flow models is the core contribution of this work. We empirically evaluate our work on continuous grid world environments with both dense and sparse rewards and find that Normalizing Flow policies converge faster and to higher rewards.

\section{Preliminaries} \label{prelims}

\xhdr{Maximum Entropy Reinforcement Learning}
The Reinforcement Learning problem can be formulated as a Markov decision process (MDP) \cite{RL} which is defined by a tuple $<\mathcal{S},\mathcal{A},p,r, \gamma>$ where $\mathcal{S}$ is the state space, $\mathcal{A}$ is the continuous action space, $p:\mathcal{S} \times \mathcal{A} \rightarrow \mathbb{P_{\mathcal{S}}}$ is the state transition distribution, $r: \mathcal{S} \times \mathcal{A} \rightarrow [r_{\min} , r_{\max} ] $ is a bounded reward emitted by the environment, and finally $\gamma$ is the discount factor. Maximum entropy RL adds an entropy term, $ \mathbb{H}(\pi(\cdot | s)) $, as a regularizer \cite{ziebart2008maximum} and leads to\cut{considers maximizing} a modified formulation of the policy gradient objective
$$
J(\pi) = \sum_{t=0}^{T} \underset{ \underset{a_t \sim \pi}{s_t \sim p} }{\E}[r(s_t,a_t) + \alpha\mathbb{H}(\pi(\cdot | s) ) ].
$$

\xhdr{Soft Actor Critic}
Soft Actor-Critic (SAC) is a policy gradient algorithm that combines several recent approaches in RL including function approximation using Neural Networks, Double Q-learning \cite{van2016deep} and entropy-regularized rewards to produce an off-policy actor-critic algorithm that is both sample efficient and emphasizes exploration \cite{haarnoja2018soft}.  


SAC learns three functions; two value functions $Q^{\pi_\theta}_\phi(s,a)$ and $V^{\pi_\theta}_\psi(s)$ playing the role of critics, and the target policy $\pi_\theta$, referred to as the actor. The $Q_\phi$ network can be learnt by minimizing a mean squared bootstrapped estimate (MSBE) while the value network, $V_\psi$, is learnt by minimizing the squared residual error.
The policy in SAC is a reparametrized Gaussian with a squashing function: $a_t = f_{\theta}(s_t;\epsilon_t) = \textnormal{tanh}(\mu_{\theta}(s_t) + \sigma_{\theta}(s_t) \cdot \epsilon_t)$ where $\epsilon_t \sim \mathcal{N}(0,I)$. Finally, the policy parameters can be learnt by maximizing the boostrapped entropy regularized reward.

\xhdr{Stochastic Computation Graphs}
Stochastic Computation Graphs \cite{StochasticCompGraph} provide a formalism for estimating the gradient of a stochastic function with learnable parameters by constructing directed acyclic graphs (DAGs). Consider for example a continuous random variable $Z$ parameterized by $\theta$, where the objective is to minimize the expected cost $L(\theta)=\mathbb{E}_{z \sim p_\theta} [f(z)]$ for some cost function $f(z)$. The gradient is defined as $\displaystyle \nabla_{\theta}L(\theta) = \nabla_{\theta} \E_{z \sim p_{\theta}}[f(z)]$. The REINFORCE estimator uses the property $\nabla_\theta p_\theta(z)  = p_\theta(z) \nabla_\theta \log( p_\theta(z))$ to re-write the gradient of the expected cost as $\nabla_\theta L(\theta) = \mathbb{E}_{z \sim p_\theta} [f(z) \nabla_\theta \log( p_\theta(z)) ]$ which we can estimate using Monte Carlo estimation \cite{williams1992simple}. 
In cases where $f$ is differentiable and $p_\theta$ can be \emph{reparameterized} through a deterministic function, $z = g(\epsilon,\theta)$, a low variance gradient estimate can be be computed by shifting the stochasticity from the distributional parameters to a standardized noise model, $\epsilon$ \cite{kingma2013auto,rezende2014stochastic}. Specifically, we can rewrite the gradient computation as $\displaystyle \nabla_{\theta}L(\theta) = \E_{\epsilon}[\nabla_g f(g(\epsilon, \theta))\nabla_{\theta}g(\epsilon,\theta)]$.
\cut{The Reparamterization trick for a set of distribution, there exits a way to generate a sample from their distribution, say $Z$, by first sampling from a simple fixed distribution $X$ and then transforming the sample using $g_\theta(x)=z$ where $z\sim Z$ \cite{kingma2013auto,rezende2014stochastic}. When such a reparameterization exists, we can rewrite $ L(\theta)= \mathbb{E}_{z \sim p_\theta} [f(z)] = \mathbb{E}_{x \sim X} [f(g_\theta(x))]$ which allows us to find the gradient rather simply:
$\nabla_\theta L(\theta)=\nabla_\theta \mathbb{E}_{x \sim X} [f(g_\theta(x))] = \mathbb{E}_{x \sim X} [\nabla_\theta f(g_\theta(x))] $. In practice, the reparametrized gradient for continuous distributions has lower variance than the \textsc{Reinforce} gradient estimator and is preferred whenever such reparametrizations exist.}

\xhdr{Normalizing Flows}
Given a parametrized density a \textit{normalizing flow} defines a sequence of invertible transformations to a more complex density over the same space via the change of variable formula for probability distributions \cite{rezende2015variational}. Starting from a sample from a base distribution, $z_0 \sim p(z)$, a mapping $f: \mathbb{R}^d \to \mathbb{R}^d$, with parameters $\theta$ that is both invertible and smooth, the log density of $z' = f(z_0)$ is defined as $\log p_{\theta}(z') = \log p(z_0) - \log \det \Big \lvert \frac{\partial f}{\partial z} \Big \rvert$.

\cut{
\begin{align}
    \log p_{\theta}(z') = \log p(z_0) - \log \det \Big \lvert \frac{\partial f}{\partial z} \Big \rvert
    \label{flow_eqn_1}
\end{align}}

Here, $p_{\theta}(z')$ is the probability of the transformed sample and $\partial f / \partial x$ is the Jacobian of $f$. To construct arbitrarily complex densities a sequence of functions, a flow, is defined and the change of density for each of these invertible transformations is applied. Thus the final sample from a flow is given by $z_k = f_K \circ f_{K-1} ... \circ f_1(z_0)$ and its corresponding density can be determined simply by $\ln p_{\theta}(z_K) = \ln p(z_0) - \sum_{k=1}^K\ln \det \Big \lvert \frac{\partial f_k}{\partial z_{k-1}} \Big \rvert$. \cut{Typically, the cost of calculating the log determinant is $O(\mathbb{D}^3)$ however through an appropriate choice of $f$ this computation cost can be brought down significantly \cite{dinh2016density}.}While there are many different choices for the transformation function, $f$, in this work we consider only RealNVP based flows as presented in \cite{dinh2016density} due to their simplicity.
\section{Method}
\xhdr{Connection with SAC}
We motivate our approach by observing the existing relationship between sampling an action in SAC and the change of variable formula for probability distributions. As outlined in Section \ref{prelims}, a sampled action in SAC is taken from a reparametrized Gaussian distribution after a squashing function is applied. The squashing function, tanh, is a bijective mapping, transforming a sample from a Gaussian distribution with infinite support to one with bounded support in $(-1,1)$. Let $u \in \mathbb{R}^D$ be the sample from the Gaussian distribution conditioned on some state $s$, then the log density of the action is given by $\log \pi(a|s) = u(a|s) - \sum_{i=1}^D \log(1-\textnormal{tanh}(u_i)^2)$, where the right term is the log determinant of the Jacobian of tanh applied component-wise. Consequently, the squashing function is a one layer Normalizing Flow.

\xhdr{Normalizing Flow Policies}
We now present how to construct Normalizing Flow policies within the SAC framework. To define the policy, $\pi(a|s)$, we first encode the state using a neural network which is then used to define the base distribution $z_0$. In principle, $z_0$ can be any probability distribution and learning can use the general REINFORCE gradient estimator. However, choosing a distribution with an available reparametrization allows for empirically lower variance gradient estimates and increased stability, a fact we verify in Section \ref{experiments}. Similar to SAC we choose $z_0$ to be a Gaussian distribution whose parameters are defined by a state-dependent neural network. Unlike SAC, we do not apply a squashing function after sampling from $z_0$, but instead apply a sequence of invertible transformations that define the flow; after which we apply the squashing function. We use a modified RealNVP style invertible transformations to define each flow layer $f_i$. Thus, the final sampled action is given by $\pi(a|s) = \textnormal{tanh}(f_k \circ ... \circ f_1(\mu_{\theta}(s) + \sigma_{\theta}(s) \cdot \epsilon)) $.
Let $u(a|s) = z_0$ be the sampled intermediate action from the reparametrized Gaussian. The corresponding log density of $a$, under the Normalizing Flow is simply $\log \pi(a|s) =  \log p(z_0) - \sum_{i=1}^k (\log \det \Big \lvert \frac{\partial f_k}{\partial z_{i-1}} \Big \rvert ) - \sum_{i=1}^D log(1-\textnormal{tanh}(z_{k_i})^2).$
\cut{
}
As the final output action has a well defined density it is easy to compute the entropy of the policy through sampling. This fact allows Normalizing Flow policies to be a convenient replacement for Gaussian policies within SAC and maximum entropy RL. 
\begin{figure}
    \includegraphics[width=1\linewidth]{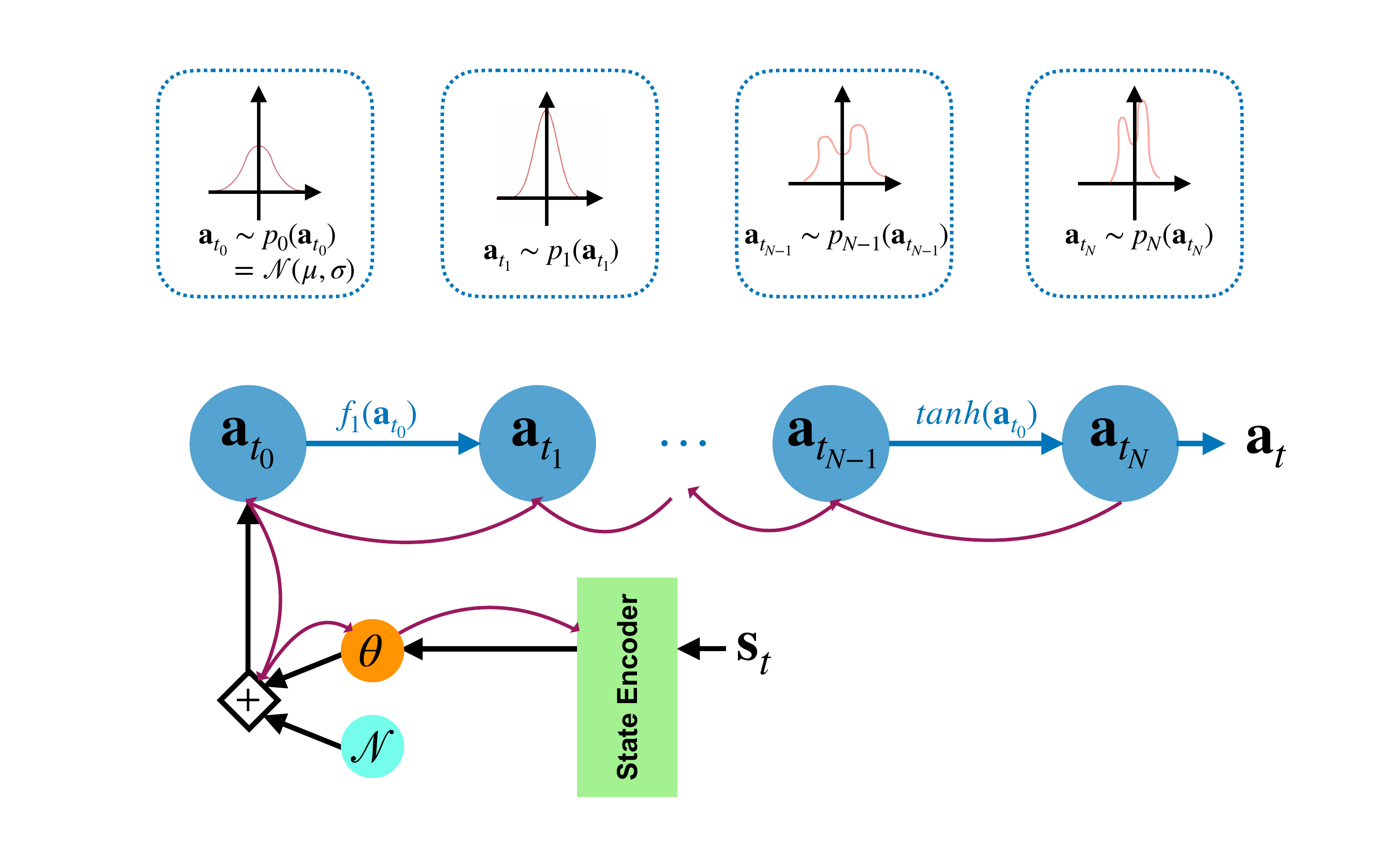}
    \vspace{-10mm}
    \caption{A sampled action from a Normalizing Flow policies}
    \label{arch_diagram}
\end{figure}

\xhdr{Stabilizing Tricks}
While the formulation above is not the only choice for defining a Normalizing Flow policy it is significantly more stable than the one presented in \cite{tang2018boosting}. We implement the state encoder as a reparametrized Gaussian which consistently leads to stable behavior after applying a few simple tricks that we now detail. We apply a series of tricks taken from the generative modeling literature to further stabilize training. Specifically, we use weight clipping for the policy parameters to enforce 1-Lipschitzness as demonstrated in Wasserstein GAN \cite{arjovsky2017wasserstein}. From a practical perspective weight clipping prevents any parameter in the policy from overflowing due to numerical precision. We hypothesize such phenomena is exacerbated in the RL setting because using a more expressive policy enables sampling extreme values before the squashing function. Furthermore, RealNVP uses a BatchNorm layer \cite{ioffe2015batch} as part of its flow. Empirically, we observe the log standard deviation of the batch can also overflow. Thus we omit the BatchNorm layer completely from our flow.
\section{Experiments} \label{experiments}
\begin{figure*}
    \includegraphics[width=0.34\linewidth]{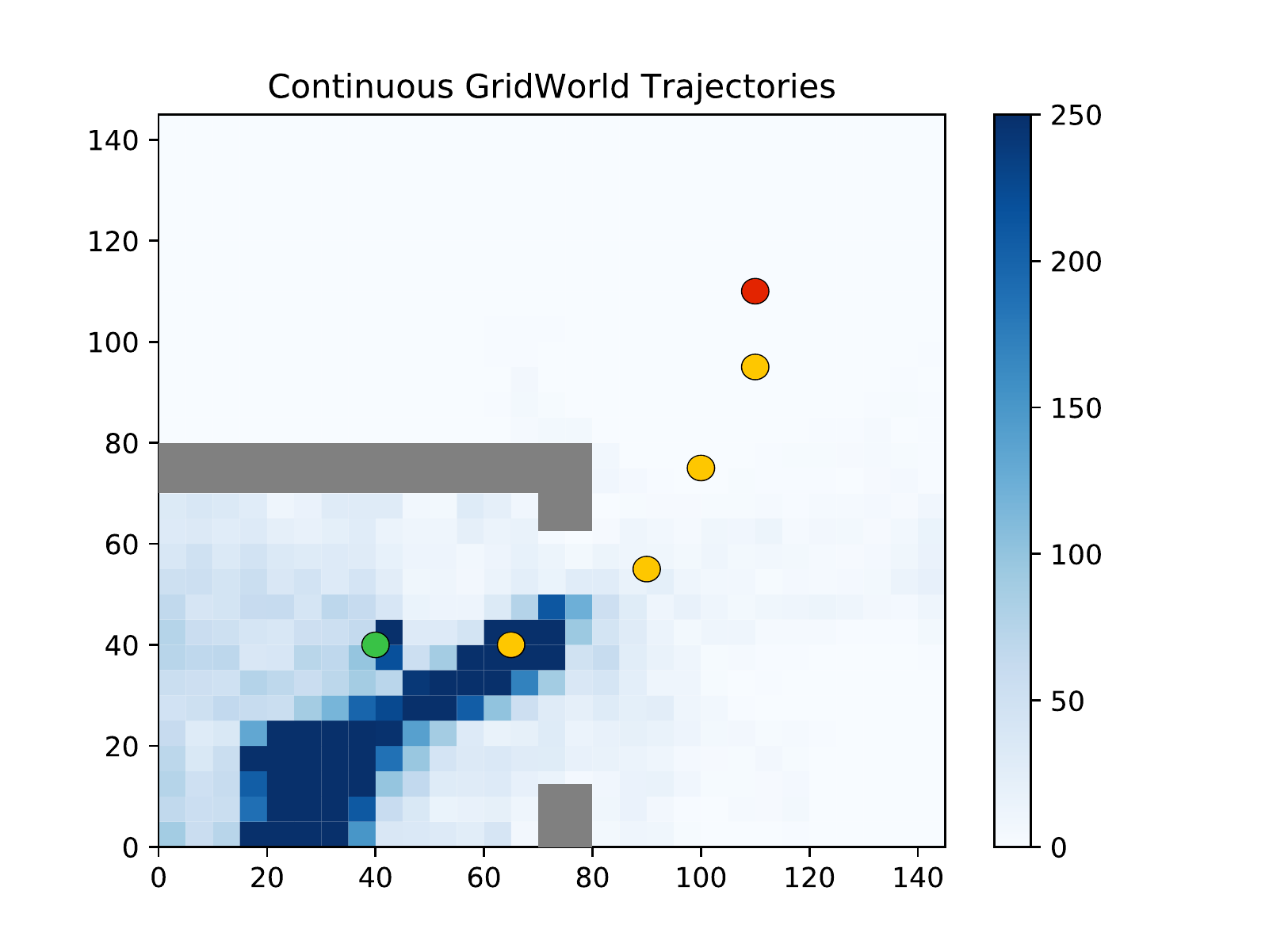}
    \includegraphics[width=0.34\linewidth]{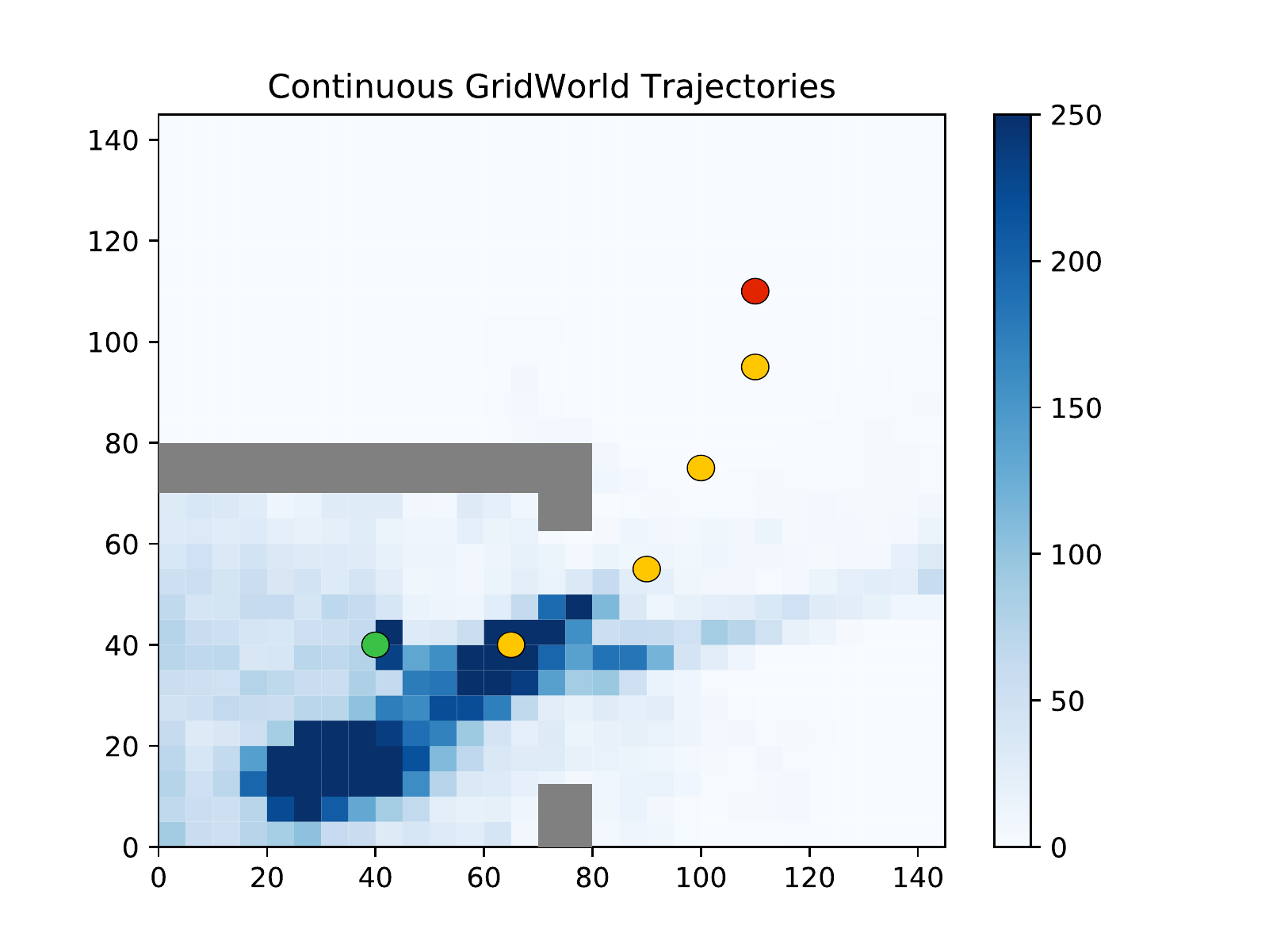}
    \includegraphics[width=0.34\linewidth]{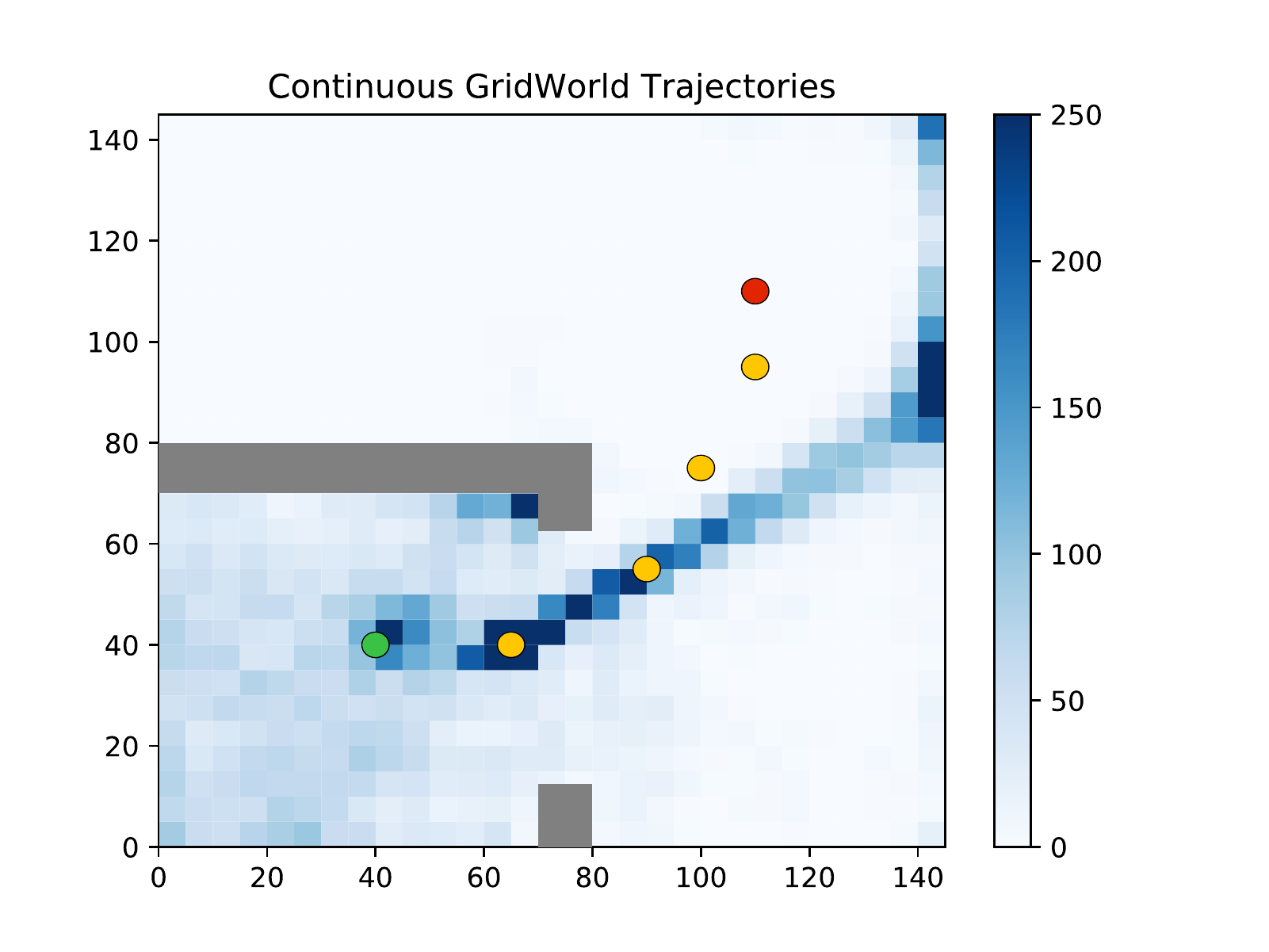}
    \includegraphics[width=0.34\linewidth]{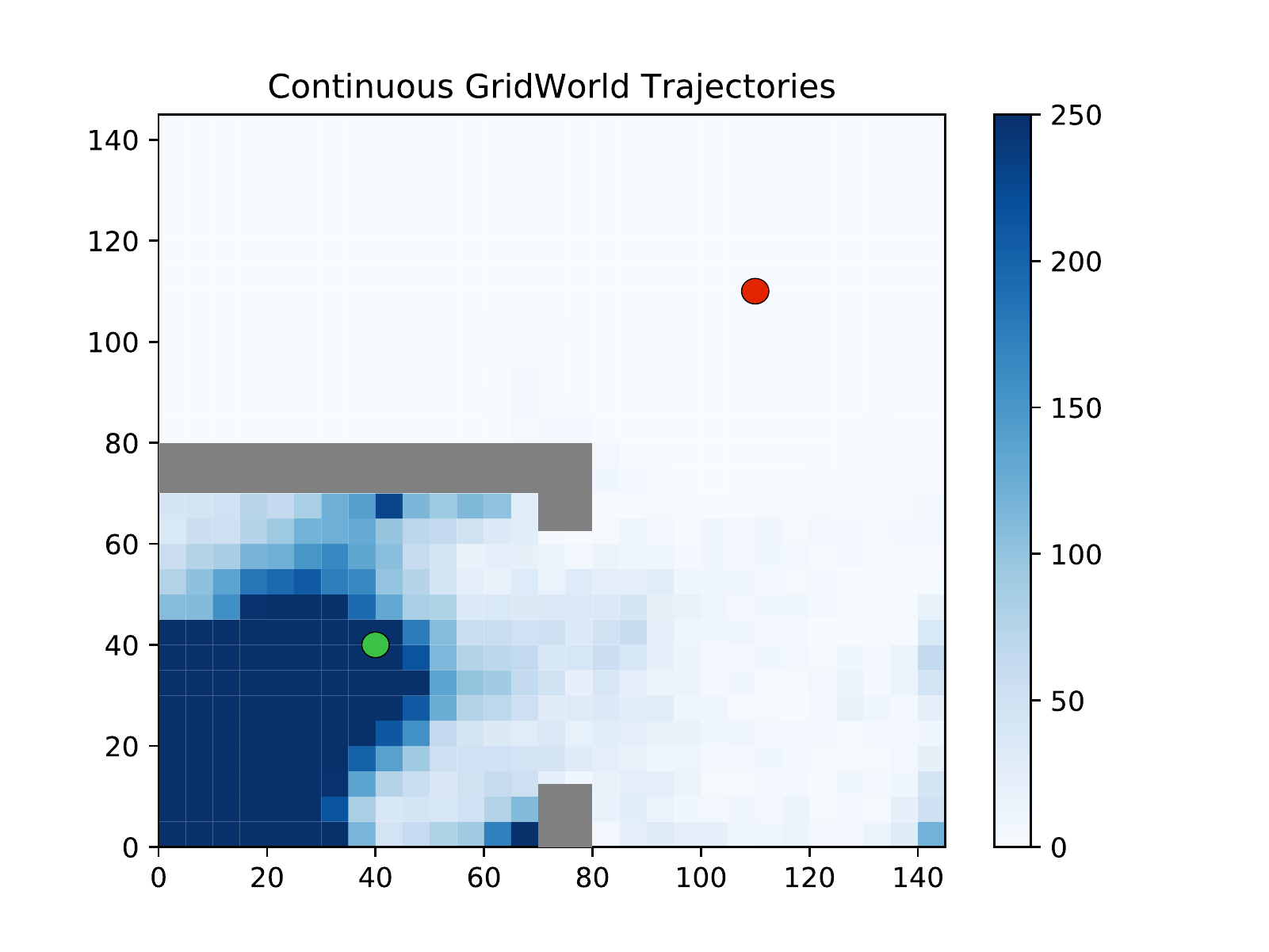}
    \includegraphics[width=0.34\linewidth]{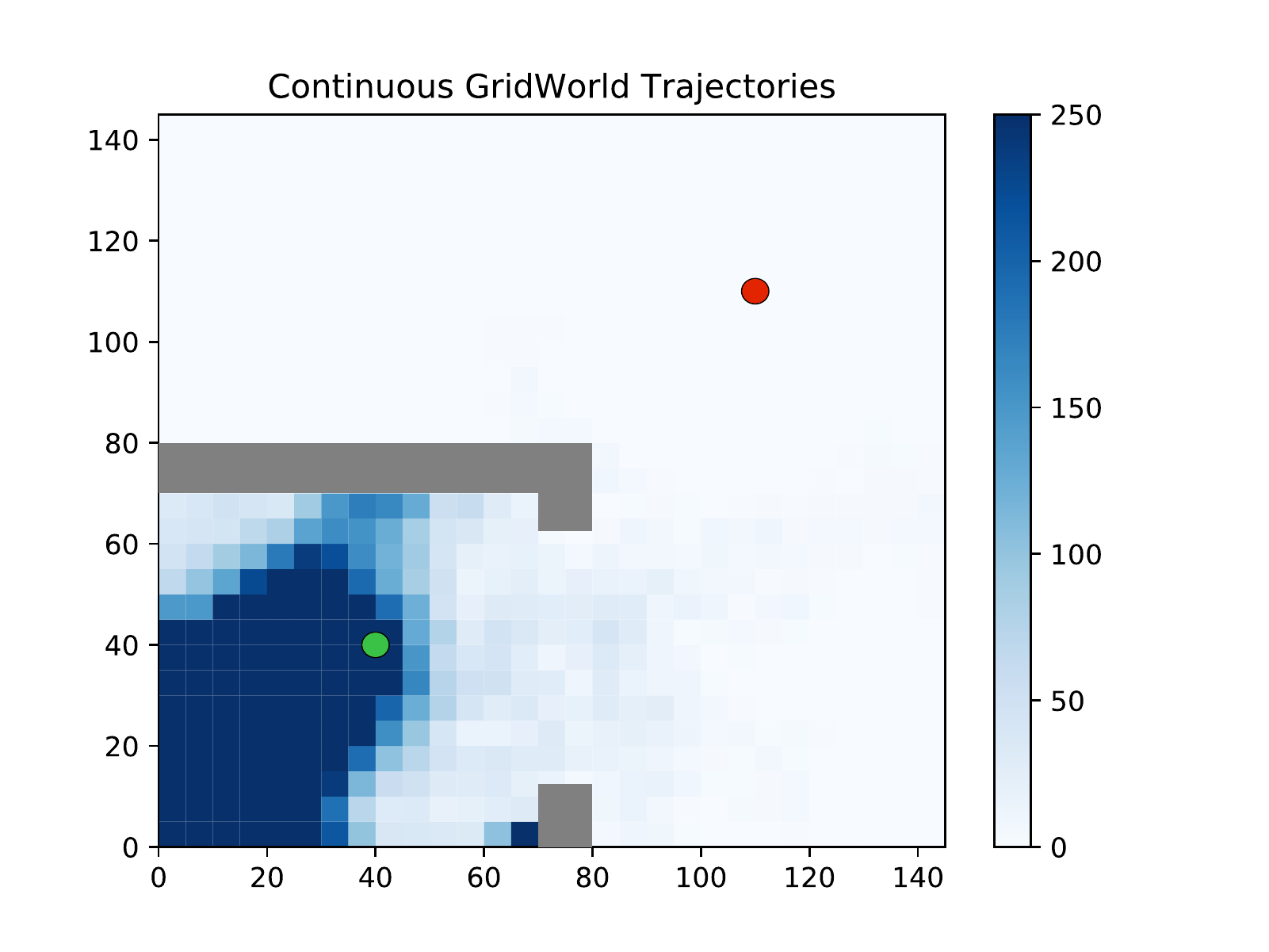}
    \includegraphics[width=0.34\linewidth]{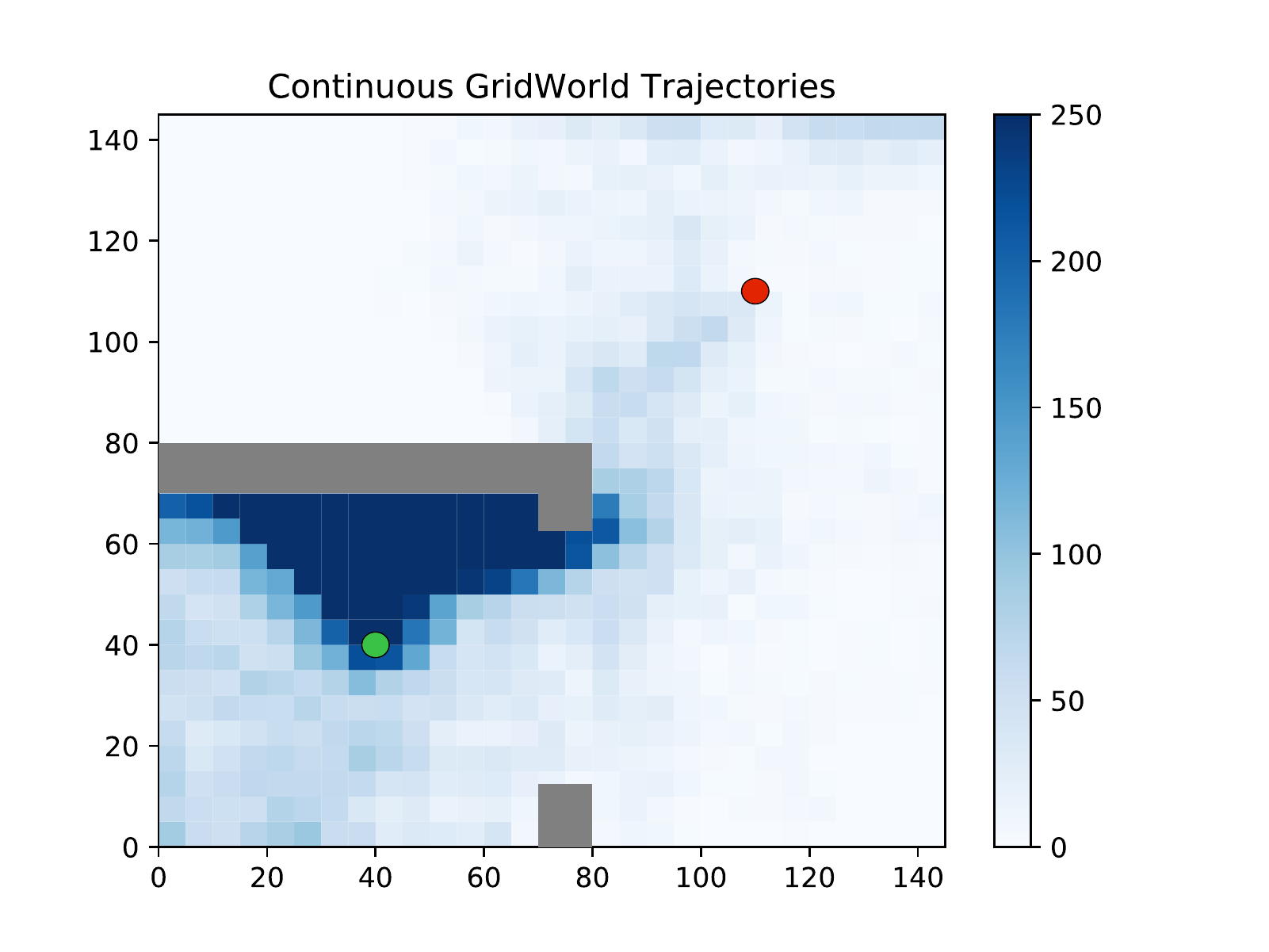}
    \vspace{-7mm}
    \caption{Heatmaps for various policies in SAC on Dense (\textbf{Top}) and Sparse (\textbf{Bottom}) reward settings. The grey areas indicate impassable walls. The start and goal states are shown as green and red points respectively. Sub-goal states are depicted as yellow points. \textbf{Left} Gaussian Policy with REINFORCE. \textbf{Middle} Gaussian Policy with Reparametrization. \textbf{Right} Normalizing Flow Policy}
    \label{smallDenseGridHeatMap}
\end{figure*}
We investigate the limitations of imposing a fixed family of probability distributions for continuous control tasks in Reinforcement Learning \footnote{Code: \url{https://github.com/joeybose/FloRL}}. To this end, we compare various implementations of SAC that modify the family of distributions used. Through our experiments we seek to answer the following questions:
\begin{itemize}
    \item[\textbf{(Q1)}] What is the impact of picking a specific family of distributions on performance?
    \item[\textbf{(Q2)}] What is the effect of different gradient estimators on the learned policy and return? 
    \item[\textbf{(Q3)}] How does exploration behavior change when using a more expressive class of policy distributions?
\end{itemize}
\xhdr{Setup and Environment} We compare variations of SAC that include modeling the policy distribution by a Normalizing Flow, a factored Gaussian (the current method used in practice) and different reparametrizable distributions such as the Exponential distribution. We also consider two variations of parameter updates; the REINFORCE update and the pathwise gradient estimator. The chosen RL environment is a continuous two-dimensional grid world introduced by \cite{oh2017value}. The goal of the continuous grid world task is for the agent, whose trajectory begins at a fixed starting point, to maximize its discounted reward for the duration of a 500 timestep episode. We experiment on two versions of this environment; a continuous grid world of size $150 \times 150$ units with sparse and dense rewards. Both environments have the same start state indicated by the green point and grey walls that the agent cannot pass as shown in Fig. \ref{smallDenseGridHeatMap}. The sparse reward environment is composed of a single goal state of value $100.0$, the red point, whereas the dense reward environment includes additional sub-goal states of value $5.0$ indicated as yellow points. These rewards are per timestep, meaning an agent can stay in a reward state until the end of the episode consisting of 500 timesteps.

\xhdr{Results} We now address questions (\textbf{Q1-Q3}).
\begin{figure}
    \includegraphics[width=0.9\linewidth]{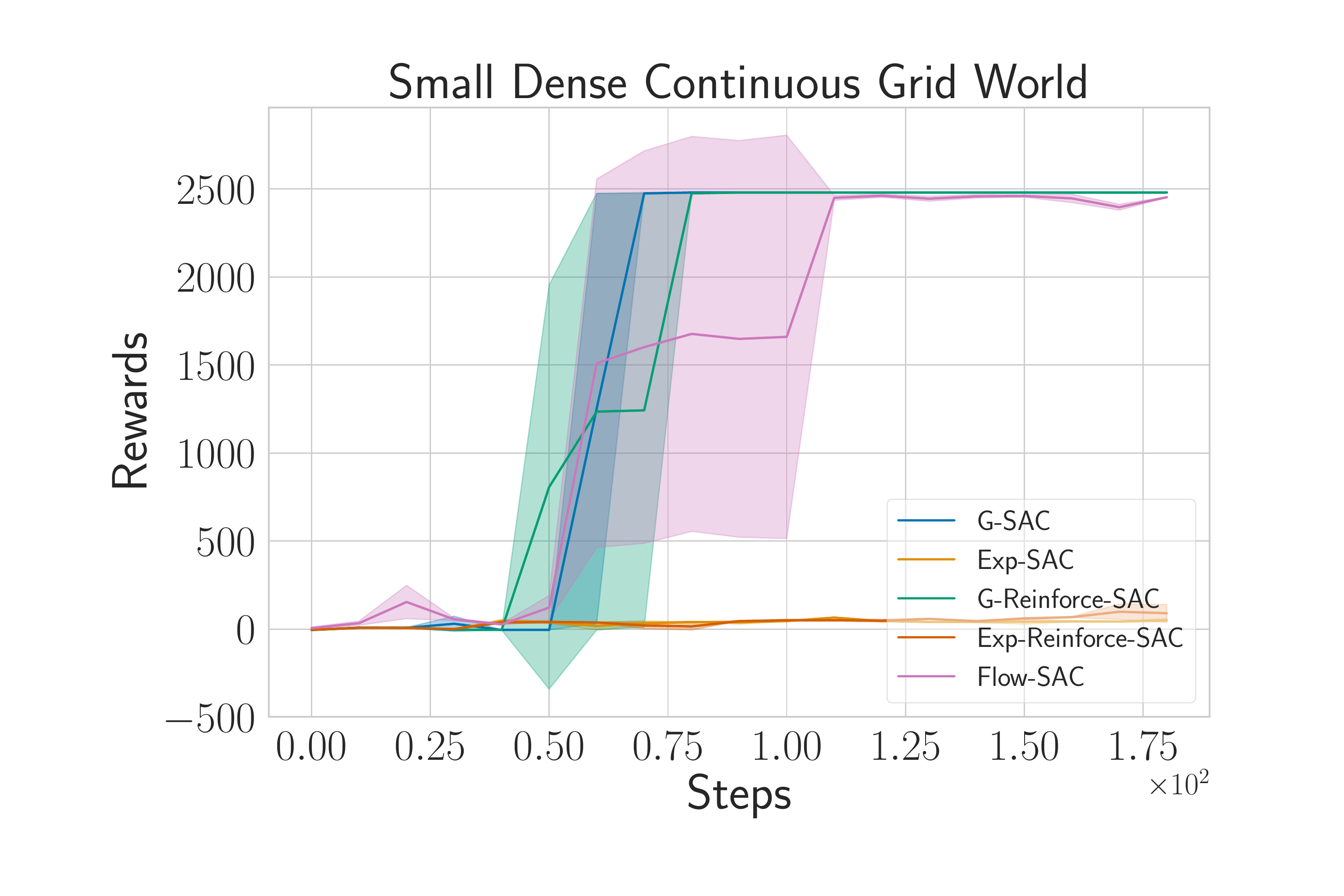}
    \includegraphics[width=0.99\linewidth]{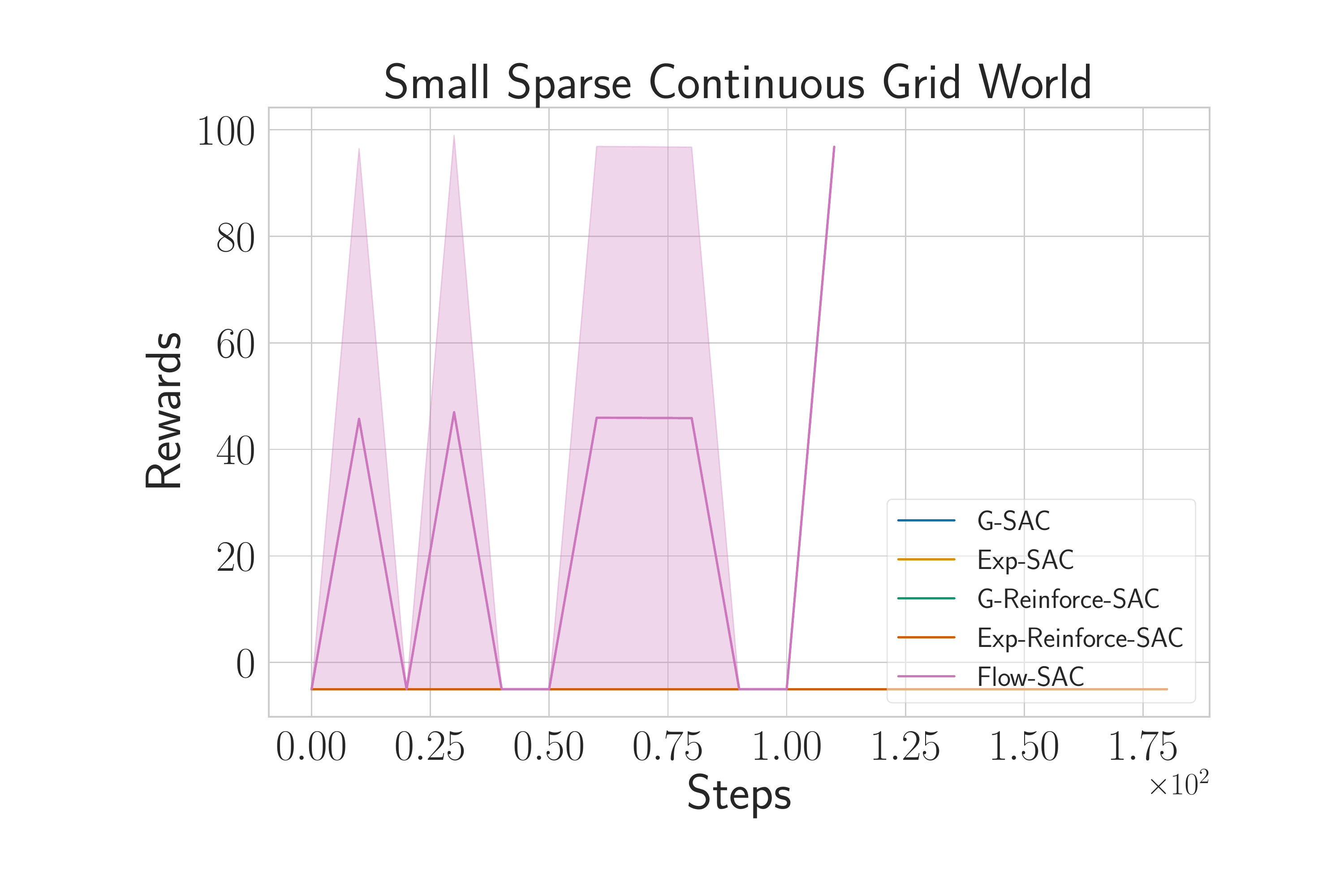}
    \vspace{-5mm}
    \caption{Performance of SAC with various policies and gradient estimators on Continuous Grid World. All runs are averaged over three runs and the shaded region represents one standard deviation.}
    \label{smallGridPerfromance}
\end{figure}

\textbf{Q1}. Empirically we observe that the choice of distribution significantly impacts model performance Fig. \ref{smallGridPerfromance}. In dense reward settings, the Gaussian policy consistently converges quickly to an optimal policy while on sparse rewards it fails to find a solution within the first $100k$ steps. The Exponential policy does significantly worse than the Gaussian and only begins to learn near the end of training, indicating the choice of distribution does matter for this task. Overall, we find that the Normalizing flows achieves a reasonable trade-off when considering both environments since it manages to get comparable performance to the Gaussian policy in the dense rewards settings but, more importantly, performs significantly better, i.e. reaching the goal state before any other distribution, when run with sparse rewards.

\textbf{Q2}. In all of our experiments, we observe that the REINFORCE gradient estimate leads to higher variance in performance Fig. \ref{smallGridPerfromance}. However both REINFORCE and reparametrized policies converge to a similar reward.

\textbf{Q3}. In Dense reward settings all three policy variations converge to the first sub-goal Fig. \ref{smallDenseGridHeatMap}. We notice the Normalizing Flow policy does not explore the environment far beyond the first and second sub-goals while both the Gaussian Policies tend to explore regions where no rewards are present. In Sparse reward settings we see that Normalizing Flow policies explore regions close to the goal state while both Gaussian policies only explore locally and far from the goal state, as substantiated in Fig. \ref{smallGridPerfromance}.
\section{Discussion and conclusion}
We introduce Normalizing Flow policies within the maximum entropy RL framework. Specifically, we replace Gaussian policies in Soft Actor Critic with modified RealNVP flows. In essence, Normalizing Flow policies have the ability to model more expressive classes of policies while maintaining the benefit of Gaussian policies, i.e. easy to sample from and having a defined probability density. We also present a few stabilizing tricks that enable training Normalizing Flow policies in the RL setting. Empirically, we observe that our approach has the ability to explore sparse reward continuous gridworld settings in a more efficient manner than Gaussian or Exponential policies, in some cases finding the optimal reward almost immediately. Additionally, our Normalizing Flow approach comes with minimal added cost on dense reward settings since our approach converges to the optimal reward at a rate similar to the Gaussian policy. In terms of directions for future work, one important design decision is the choice of architecture when defining the Normalizing Flow. In this work, we considered Flows based on the RealNVP architecture but applying more recent variants such as Glow \cite{kingma2018glow}, NAF \cite{huang2018neural}, and FFJORD \cite{grathwohl2018ffjord} to the RL setting is a natural direction for future work.

\clearpage
\bibliography{bibliography.bib}

\begin{thebibliography}{21}
\providecommand{\natexlab}[1]{#1}
\providecommand{\url}[1]{\texttt{#1}}
\expandafter\ifx\csname urlstyle\endcsname\relax
  \providecommand{\doi}[1]{doi: #1}\else
  \providecommand{\doi}{doi: \begingroup \urlstyle{rm}\Url}\fi

\bibitem[Arjovsky et~al.(2017)Arjovsky, Chintala, and
  Bottou]{arjovsky2017wasserstein}
Arjovsky, M., Chintala, S., and Bottou, L.
\newblock Wasserstein gan.
\newblock \emph{arXiv preprint arXiv:1701.07875}, 2017.

\bibitem[Chou et~al.(2017)Chou, Maturana, and Scherer]{chou2017improving}
Chou, P.-W., Maturana, D., and Scherer, S.
\newblock Improving stochastic policy gradients in continuous control with deep
  reinforcement learning using the beta distribution.
\newblock In \emph{Proceedings of the 34th International Conference on Machine
  Learning-Volume 70}, pp.\  834--843. JMLR. org, 2017.

\bibitem[Dinh et~al.(2016)Dinh, Sohl-Dickstein, and Bengio]{dinh2016density}
Dinh, L., Sohl-Dickstein, J., and Bengio, S.
\newblock Density estimation using real nvp.
\newblock \emph{arXiv preprint arXiv:1605.08803}, 2016.

\bibitem[Grathwohl et~al.(2018)Grathwohl, Chen, Betterncourt, Sutskever, and
  Duvenaud]{grathwohl2018ffjord}
Grathwohl, W., Chen, R.~T., Betterncourt, J., Sutskever, I., and Duvenaud, D.
\newblock Ffjord: Free-form continuous dynamics for scalable reversible
  generative models.
\newblock \emph{arXiv preprint arXiv:1810.01367}, 2018.

\bibitem[Haarnoja et~al.(2018)Haarnoja, Zhou, Abbeel, and
  Levine]{haarnoja2018soft}
Haarnoja, T., Zhou, A., Abbeel, P., and Levine, S.
\newblock Soft actor-critic: Off-policy maximum entropy deep reinforcement
  learning with a stochastic actor.
\newblock \emph{arXiv preprint arXiv:1801.01290}, 2018.

\bibitem[Heess et~al.(2015)Heess, Wayne, Silver, Lillicrap, Erez, and
  Tassa]{heess2015learning}
Heess, N., Wayne, G., Silver, D., Lillicrap, T., Erez, T., and Tassa, Y.
\newblock Learning continuous control policies by stochastic value gradients.
\newblock In \emph{Advances in Neural Information Processing Systems}, pp.\
  2944--2952, 2015.

\bibitem[Huang et~al.(2018)Huang, Krueger, Lacoste, and
  Courville]{huang2018neural}
Huang, C.-W., Krueger, D., Lacoste, A., and Courville, A.
\newblock Neural autoregressive flows.
\newblock \emph{arXiv preprint arXiv:1804.00779}, 2018.

\bibitem[Ioffe \& Szegedy(2015)Ioffe and Szegedy]{ioffe2015batch}
Ioffe, S. and Szegedy, C.
\newblock Batch normalization: Accelerating deep network training by reducing
  internal covariate shift.
\newblock \emph{arXiv preprint arXiv:1502.03167}, 2015.

\bibitem[John~Schulman(2015)]{StochasticCompGraph}
John~Schulman, Nicolas~Heess, T. W. P.~A.
\newblock Gradient estimation using stochastic computation graphs.
\newblock In \emph{Advances in Neural Information Processing Systems}, pp.\
  3528--3536, 2015.

\bibitem[Kingma \& Dhariwal(2018)Kingma and Dhariwal]{kingma2018glow}
Kingma, D.~P. and Dhariwal, P.
\newblock Glow: Generative flow with invertible 1x1 convolutions.
\newblock In \emph{Advances in Neural Information Processing Systems}, pp.\
  10215--10224, 2018.

\bibitem[Kingma \& Welling(2013)Kingma and Welling]{kingma2013auto}
Kingma, D.~P. and Welling, M.
\newblock Auto-encoding variational bayes.
\newblock \emph{arXiv preprint arXiv:1312.6114}, 2013.

\bibitem[Oh et~al.(2017)Oh, Singh, and Lee]{oh2017value}
Oh, J., Singh, S., and Lee, H.
\newblock Value prediction network.
\newblock In \emph{Advances in Neural Information Processing Systems}, pp.\
  6118--6128, 2017.

\bibitem[Rezende \& Mohamed(2015)Rezende and Mohamed]{rezende2015variational}
Rezende, D.~J. and Mohamed, S.
\newblock Variational inference with normalizing flows.
\newblock \emph{arXiv preprint arXiv:1505.05770}, 2015.

\bibitem[Rezende et~al.(2014)Rezende, Mohamed, and
  Wierstra]{rezende2014stochastic}
Rezende, D.~J., Mohamed, S., and Wierstra, D.
\newblock Stochastic backpropagation and approximate inference in deep
  generative models.
\newblock \emph{31st International Conference on Machine Learning}, 2014.

\bibitem[Schulman et~al.(2015)Schulman, Heess, Weber, and
  Abbeel]{schulman2015gradient}
Schulman, J., Heess, N., Weber, T., and Abbeel, P.
\newblock Gradient estimation using stochastic computation graphs.
\newblock In \emph{Advances in Neural Information Processing Systems}, pp.\
  3528--3536, 2015.

\bibitem[Sutton \& Barto(1998)Sutton and Barto]{RL}
Sutton, R.~S. and Barto, A.~G.
\newblock \emph{Reinforcement learning - an introduction}.
\newblock Adaptive computation and machine learning. {MIT} Press, 1998.
\newblock ISBN 0262193981.
\newblock URL \url{http://www.worldcat.org/oclc/37293240}.

\bibitem[Tang \& Agrawal(2018)Tang and Agrawal]{tang2018boosting}
Tang, Y. and Agrawal, S.
\newblock Boosting trust region policy optimization by normalizing flows
  policy.
\newblock \emph{arXiv preprint arXiv:1809.10326}, 2018.

\bibitem[Van~Hasselt et~al.(2016)Van~Hasselt, Guez, and Silver]{van2016deep}
Van~Hasselt, H., Guez, A., and Silver, D.
\newblock Deep reinforcement learning with double q-learning.
\newblock In \emph{Thirtieth AAAI Conference on Artificial Intelligence}, 2016.

\bibitem[Williams(1992)]{williams1992simple}
Williams, R.~J.
\newblock Simple statistical gradient-following algorithms for connectionist
  reinforcement learning.
\newblock \emph{Machine learning}, 1992.

\bibitem[Ziebart(2010)]{ziebart2010modeling}
Ziebart, B.~D.
\newblock Modeling purposeful adaptive behavior with the principle of maximum
  causal entropy.
\newblock 2010.

\bibitem[Ziebart et~al.(2008)Ziebart, Maas, Bagnell, and
  Dey]{ziebart2008maximum}
Ziebart, B.~D., Maas, A.~L., Bagnell, J.~A., and Dey, A.~K.
\newblock Maximum entropy inverse reinforcement learning.
\newblock In \emph{Aaai}, volume~8, pp.\  1433--1438. Chicago, IL, USA, 2008.

\end{thebibliography}
\bibliographystyle{icml2019}

\end{document}